\RequirePackage{fix-cm}
\documentclass[smallextended]{svjour3}       
\smartqed  
\usepackage{graphicx}
\usepackage{subfig}

\usepackage{framed,multirow}

\usepackage{amssymb}
\usepackage{latexsym}
\usepackage{subfig}

\begin{document}

\title{Anomaly detecting and ranking of the cloud computing platform by multi-view learning
}

\author{Jing Zhang  }


\institute{F. Author \at
              School of computer and information technology, Liaoning Normal University, Liushu south street NO.1 Ganjingzi district, Dalian and 116018, China \\
              Tel.: +86-0411-85992418\\
              Fax: +86-0411-85992418\\
              \email{zhangjing\_0412@163.com}           
}

\date{Received: date / Accepted: date}

\maketitle

\begin{abstract}
Anomaly detecting as an important technical in cloud computing is applied to support smooth running of the cloud platform. Traditional detecting methods based on statistic, analysis, etc. lead to the high false-alarm rate due to non-adaptive and sensitive parameters setting. We presented an online model for anomaly detecting using machine learning theory. However, most existing methods based on machine learning linked all features from difference sub-systems into a long feature vector directly, which is difficult to both exploit the complement information between sub-systems and ignore multi-view features enhancing the classification performance. Aiming to this problem, the proposed method automatic fuses multi-view features and optimize the discriminative model to enhance the accuracy. This model takes advantage of extreme learning machine (ELM) to improve detection efficiency. ELM is the single hidden layer neural network, which is transforming iterative solution of the output weights to solution of linear equations and avoiding the local optimal solution. Moreover, we rank anomies according to the relationship between samples and the classification boundary, and then assigning weights for ranked anomalies, retraining the classification model finally. Our method exploits the complement information between sub-systems sufficiently, and avoids the influence from imbalance dataset, therefore, deal with various challenges from the cloud computing platform. We deploy the privately cloud platform by Openstack, verifying the proposed model and comparing results to the state-of-the-art methods with better efficiency and simplicity.
\end{abstract}

\section{Introduction}
\label{intro}
Cloud computing makes possible auto-scaling and using resources at all time, moreover, avoid waste or out of expectation when developers deploy applications[1, 2]. However, multiple anomalies in the cloud platform become both the major bottleneck for high available and the primary cause to impede development, it is important to how to build efficient anomaly detecting models. Previous studies in anomaly detecting focus on statistical analysis of running curve, which is computing and comparing curves and setting the threshold values to find anomalies by operators. The kind of methods produces both low accuracy and the high false alarm ratio due to adjust parameters by manually in unknown data distribution. Therefore, the anomaly detecting problem in cloud computing is often formulated as a data association problem based on machine learning. Machine learning models that contains data drive, non-linear fitting and incremental provide preconditions for the problem of detection defining and solving.
\par In detecting-by-learning techniques, the performance depends on ten detection accuracy and the challenge from the high-dimension vector that consists of multiple sub-systems [3]. Information redundancy and noisy from different features will result in failures of detecting. Therefore, feature extraction and dimensionality reduction as efficient means enhance the detecting performance in cloud computing based on virtualization. Sub-space learning extracts principal features by building the optimum projection space such as principal component analysis (PCA) [4], locality preserving projections (LPP) [5], linear discriminant analysis (LDA) [6]. In order to improve expression of features, independent component analysis (ICA) [7] is proposed. Robust dimensionality reduction methods by sparse representation and low-rank learning including robust PCA model [8], sub-space recover model [9], etc. Above methods are applied in information selection in cloud computing due to both high-integrity date description and high-effective computing. Guan, etc. exploit PCA to obtain the most relevant components, and then adaptive kalman filter (AKF) regard as classifier to enhance the detecting performance [10]. Fu, etc., utilize PAC to extract features of the cloud platform, which is different to the research from Guan, and it is selecting features in indicator vectors by mutual information method, and then extracting the most important features on selected features based on PCA [11]. Lan, etc., analyze in such a way as to distinguish between PCA and ICA in cloud computing detecting, and accept that ICA achieves higher and faster performance in experiments [12]. However, above sub-space methods only contain the global structure and ignore the local structure in original datasets. Local projection remaining model achieves more accurate describe of data [13, 14]. In order to solve the problem that linear method induces high computational complexity and limited expression original information, Fujimaki, etc., rebuild the state space by inducing kernel model [15]. Faschi, etc., achieve the feature extraction model based on regression metric analysis, which is used to find anomaly in cloud platform [16]. However, many of these methods are limited due to all of features from sub-systems are linked to a long feature vector ignoring the complementary information from difference sub-systems. Moreover, above methods does not consider to optimize the discriminative model to enhance the performance.
Finding effective and robust learning models relies on appropriate discriminative model building processing of the anomaly detecting. Deng, etc. utilize SMO network to predict anomaly in Iaas cloud platform obtaining accurate results [17]. Sauvanaud, etc. exploit the dynamic index set both to update anomaly class and to compute the classification of the mean value adapting the changing of data distribution [18]. Fu, etc. first, describe the normal of the system based on Bayesian model; second, confirm the anomaly and obtain labeled samples; finally, semi-supervised model based on decision trees is used to predict anomaly in future [19]. Lan, etc. proposed the automatic recognition method for large scale distributed systems, which is using unsupervised model to detect abnormal nodes increasing the performance [20]. In order to reduce complexity of the algorithm, Wang, etc. find K-nearest nodes based on R-tree index method [21]. Unsupervised models overcome the difficult that samples tag labels by manual and achieve better performance in various of application fields [22-34], but these models high dependent on the distribution of samples. Moreover, it is an important influence how to design the metric method. Therefore, anomaly detecting methods by supervised have gained widespread concern. Beak, etc. tag samples and then utilize classification model to find anomaly [35]. Wang, etc. utilize the entropy model transforming samples into time series to enhance the accuracy [36]. Tan, etc. build on anomaly early-warning system by mixing between the Markov Model and Enhanced Bayesian Networks [37]. Yao, etc. proposed the accompanied detection model based on C4.5 classification model, which is defining both log-primary from all of samples and log-accompanied from abnormal samples. In this method, log-primary are used to train the classification model, and log-accompanied are used to recognize the type of anomalies [38]. Liu, etc. exploit SMO network to achieve automatic recognition and detection [39]. However, above methods may not be sufficiently for the definition of detecting-by-learning. For this reason, first, ignore the complementary information from multiple sub-systems. Moreover, divide into two single steps including extraction feature and classification, which is resulting from the failure supervised information in extraction feature processing. Second, the data distribution is imbalance from the cloud platform, which is reducing the detecting accuracy. Third, it is different from the traditional classification problem of anomaly detecting is the first sequence steps of anomaly handing, and the most suitable handing way is used to approach anomaly ranked. Therefore, the target that the anomaly detecting problem is defined as data associate based on detecting-by-learning is obtain anomaly set and anomaly ranking by learning models.
Overview of our algorithm is illustrated in Figure 1. We pose anomaly detecting as a data learning problem, which is solved by multi-view learning model. Our method achieves both automatic fuses multi-view features from multiple sub-systems of the cloud platform and obtains optimized discriminative model by improved extreme learning machine. In order to handle anomaly with distinguished methods and consider imbalance problem, we proposed the novel method to rank the set of anomaly, and then using ranking results to optimize the classification model to enhance robust under imbalance distribution. The proposed model based on ELM has the following characteristics:
\begin{enumerate}
  \item We provide an online method to detect anomaly by multi-view learning based on ELM without manual intervention.
  \item The proposed model achieves that multi-view features automatic fuse from multiple sub-systems according to supervised information by iterating to minimize the train error, which is exploit the complementary information substantially and obtain the optimal solution space under currently features.
  \item Ranking the set of anomaly by proposing the novel model for post-processing, and then generate weight to retrain the classification model to enhance the robustness for imbalance distribution.
  \item Through the proposed model by learning, we manage various challenges from high-speed data stream, high-dimension index set, imbalance distribution anomaly, and so on.
\end{enumerate}
For the rest of this paper, we introduce ELM in section 2. In section 3 and section 4, we proposed multi-view model to obtain the set of anomaly, ranking anomaly and optimal the classification model by means of adapting weights from ranking results. In section 5, we utilize collected data from the private cloud platform to evaluate the proposed method and comparing it with existing detecting techniques.
\section{Preliminaries: ELM and OSELM}
\subsection{ELM}
In order to facilitate the understanding of our method, this section briefly reviews the related concepts and theories of ELM and developed OSELM.
\par Extreme learning machine is improved by single hidden layer neural network (SLFNs): assume given $N$ samples $(X,T)$, where $X=[x_{1},x_{2},...,x_{N}]^{T} \in\mathbb{R}^{d\times N}$, $T=[t_{1},t_{2},...,t_{N}]^{T} \in\mathbb{R}^{\tilde{N}\times N}$, and $t_{i}=[t_{i1},t_{i2},...,t_{im}]^{T} \in\mathbb{R}^{m}$. The method is used to solve multi-classification problems, and thereby the number of network output nodes is $m(m\geq 2)$. There are $\widetilde{N}$ hidden layer nodes in networks, and activation function $h(\cdot)$ can be Sigmoid or RBF: $\sum_{i=1}^{\widetilde{N}} \beta_{i}h(a_{i}x_{j}+b_{j})=o_{j}$
where $j=1,\cdots,\tilde{N}$, $a_{j}=[a_{j1},a_{j2},\cdots,a_{jd}]^{T}$ is the input weight vector, and $\beta_{j}=[\beta_{j1},\beta_{j2},...,\beta_{jm}]^{T}$ is the output weight vector. Moreover, $a_{j}$, $b_{j}$ can be generated randomly, which is known by. Written in matrix form: $H\beta =T$, where $H_{i} = [h_{1}(a_{1}x_{1}+b_{1}),\cdots, h_{N}(a_{\widetilde{N}}x_{1}+b_{\widetilde{N}})]$. Moreover, the solution form of $H\beta=T$ can be written as: $\hat{\beta}=H^{\dag}T$, where $H^{\dag}$ is the generalized inverse matrix of $H$. ELM minimize both the training errors and the output weights. The expression can be formulated based on optimization of ELM:
\begin{equation}
\begin{array}{c}
  \rm Minimize: \it \frac{1}{2}\|\beta\|^{2}_{2}+C\frac{1}{2}\sum_{i=1}^{N} \|\xi_{i}\|^{2}_{2}\\
  \\
  \rm Subject \; to: \it t_{i} \beta\cdot h(x_{i})\geq 1-\xi_{i},i=1,...,N\\
  \\
  \xi_{i}\geq 0,i=1,...,N
\end{array}
\end{equation}
where $\xi_{i}=\left(
                 \begin{array}{ccc}
                   \xi_{i,1} & \cdots & \xi_{i,m} \\
                 \end{array}
               \right)
$ is the vector of the training errors. We can solve the above equation based on KKT theory by Lagrange multiplier, and can obtain the analytical expression of the output weight: $\hat{\beta} = H^{T}(\frac{I}{C}+HH^{T})^{-1}T$. The output function of ELM is: $f(x)=h(x)\hat{\beta}=h(x)H^{T}(\frac{I}{C}+HH^{T})^{-1}T$.
\subsection{OSELM}
The above model is used to solve classification problem for static batch data. Aiming to this problem Rong et al. proposed an increment classification model OSELM [51]. It is an online solving algorithm based on ELM. The model trains the $\Delta N(\Delta N\geq 1)$ chunk of new samples to obtain new model, then uses matrix calculation with the original model. Through the above calculation, the new output weight matrix
$\hat{\beta} _{N + \Delta N}$ is obtained. When the new $\Delta N$ chunk arrives, the hidden output weight matrix is updated. The expression is listed as follows:
\[{H_{N + \Delta N}} = \left[ {\begin{array}{*{20}{c}}
{h({x_1};{a_1},{b_1})}& \cdots &{h({x_1};{a_{\tilde N}},{b_{\tilde N}})}\\
 \vdots &{}& \vdots \\
{h({x_N};{a_1},{b_1})}& \cdots &{h({x_N};{a_{\tilde N}},{b_{\tilde N}})}\\
{h({x_{N + 1}};{a_1},{b_1})}& \cdots &{h({x_{N + 1}};{a_{\tilde N}},{b_{\tilde N}})}\\
 \vdots &{}& \vdots \\
{h({x_{N + \Delta N}};{a_1},{b_1})}& \cdots &{h({x_{N + \Delta N}};{a_{\tilde N}},{b_{\tilde N}})}
\end{array}} \right] = \left[ {\begin{array}{*{20}{c}}
{{h_1}}\\
 \vdots \\
{{h_N}}\\
{{h_{N + 1}}}\\
 \vdots \\
{{h_{N + \Delta N}}}
\end{array}} \right] = \left[ {\begin{array}{*{20}{c}}
{{H_N}}\\
{{H_{\Delta N}}}
\end{array}} \right]\]
where
${h_{N + k}} = {\left[ {\begin{array}{*{20}{c}}
{h({x_{N + k}};{a_1},{b_1})}& \cdots &{h({x_{N + k}};{a_{\tilde N}},{b_{\tilde N}})}
\end{array}} \right]^T}(k = 1, \ldots ,\Delta N)$ is the $k$th new sample corresponding the vector. Therefore, the output vector is
${T_{\Delta N}} = {\left[ {\begin{array}{*{20}{c}}
{t_{N + 1}^{}}& \cdots &{t_{N + \Delta N}^{}}
\end{array}} \right]^T}$.
Therefore, the incremental expression of the output weight is obtained:
\begin{equation}
{\hat{\beta} _{N + \Delta N}} = {(H_N^T{H_N} + H_{\Delta N}^T{H_{\Delta N}})^{ - 1}}(H_N^T{T_N} + H_{\Delta N}^T{T_{\Delta N}})
\end{equation}
Let ${G_0} = {(H_N^T{H_N})^{ - 1}}$, and the incremental expression $G_1$ can be written as:
\begin{equation}
{G_1}^{ - 1} = {G_0}^{ - 1} + H_{\Delta N}^T{H_{\Delta N}}
\end{equation}
\par According to the equation (4) and (5), the new output weight matrix $\hat{\beta} _{N + \Delta N}$ becomes:
\begin{equation}
{\hat{\beta} _{N + \Delta N}} = {G_1}({H_N}{T_N} + H_{\Delta N}^T{T_{\Delta N}}) = {\beta _N} + {G_1}H_{\Delta N}^T({T_{\Delta N}} - {H_{\Delta N}}{\beta _N})
\end{equation}
where ${G_1} = {(G_0^{ - 1} + H_{\Delta N}^T{H_{\Delta N}})^{ - 1}}$.
\par According to the above equation, we formulate the further expression:
\begin{equation}
{G_1} = {G_0} - {G_0}H_{\Delta N}^T{({I_{\Delta N}} + {H_{\Delta N}}{G_0}H_{\Delta N}^T)^{ - 1}}{H_{\Delta N}}{G_0}
\end{equation}
From the above learning process, OSELM trains new model by adjusting the original model according to the equation (7) when the new dynamic samples are arriving.
\section{The proposed anomaly detecting and ranking model}
\label{sec:1}
In order to obtain anomaly in real-time from data stream of the cloud computing platform, we proposed an incremental detecting model based on multi-view features, ranking abnormal samples that is prerequisite of anomaly handling generate weights to feedback adjustment the classification model at current to enhance the robustness for imbalance samples of anomaly. The workflow of proposed model is show in Fig. 2., and it divides into two parts: local training and online training. In local training processing, the proposed multi-view features model is used to automatic fuse difference features and achieve optimized discriminative model. In online training processing, first, fuse multiple features according to the local learning structure, which reduce time consuming from retraining all of samples including local and online samples. Second, rank anomalies detected to handle differently. Finally, set self-adapting weights for anomalies that are used to adjust the classification model to avoid the influence of imbalance distribution.

\subsection{Multi-view features fusion and discriminative optimization}
\label{sec:2}
The number of state indicators from difference subsystems in the cloud computing platform belongs to the range from dozens to hundreds, which is composed the high-dimensional feature space. However, the information and noisy between sub-systems will influence the detected accuracy in subsequent calculations and reduce the performance of detecting. Most traditional models link multiple features into the long vector, and then extract principal components from this vector to avoid information redundancy. It is difficult to mining potential and complementary information from multiple features. Aiming above problem, we automatic fuse multiple features and optimize classification model by iterative solving. The proposed method based on ELM that is used to classify samples. However, when the sample contains various features, the ELM model is difficult applied to solve multiple features of the same sample. In this paper, the the proposed fusion method based on ELM is describe as follow:
\par Given $N$ the different samples, which contain $V$ features of each sample collected in multiple ways. The feature $v$ corresponds to the samples are: $(x_i^{(v)},t_i^{(v)})$, where $ {X^{(v)}} = {\left[ {x_1^{(v)}, \ldots ,x_N^{(v)}} \right]^T} \in\mathbb{R}{^{{D} \times N}}$, $t_i^{(v)} = {\left[ {t_{i1}^{(v)}, \ldots ,t_{im}^{(v)}} \right]^T} \in\mathbb{R}{^m}$. Meanwhile, the same sample corresponds to the same class, then there is: $t_i^{(1)} = t_i^{(2)} = t_i^{(V)}$. The output weight $\beta$ is solved by using the samples that contain combined multi-view features. The optimization equation is as follows:
\begin{equation}
\begin{array}{l}
Minimize: \frac{1}{2}\left\| \beta  \right\|_2^2 + \sum\limits_{v = 1}^V {{k^{\left( v \right)}}} \left( {\sum\limits_{j = 1}^N {\left\| {\varepsilon _j^{(v)}} \right\|_2^2} } \right)\\
Subject \; to: \begin{array}{*{20}{c}}
{}&{\left( {k \cdot {H^{(v)}}} \right) \cdot \beta }
\end{array} = t_i^T - {\left( {\varepsilon _i^{(v)}} \right)^T}\\
\begin{array}{*{20}{c}}
{}&{}&{\sum\limits_{v = 1}^V {{k^{\left( v \right)}}}  = 1\begin{array}{*{20}{c}}
{}&{k > 0}
\end{array}}
\end{array}
\end{array}
\end{equation}
where $k_{j}$ is the combined parameter that corresponds to the single-features. $\xi _i^{(v)} = \left( {\xi _{i1}^{(v)}, \ldots ,\xi _{im}^{(v)}} \right)$ is training error vector that corresponds to feature $v$. $\beta$ is the output weight vector for different feature space in the equation (2). According to the equation (2), the target is to obtain the minimum value of the training error that combine the different feature space. However, according to the equation, the solution of $[{k_1},{k_2}]$ may be (0,1)/(1,0). In this situation, it will degenerates to the single feature model, and other features are failed. Therefore, we introduction the high power factor $r$, and define $r \ge 2$. The equation is improved as follows:
\begin{equation}
\begin{array}{l}
Minimize: \frac{1}{2}\left\| \beta  \right\|_2^2 + \sum\limits_{v = 1}^V {{k^{r{\left( v \right)}}}} \left( {\sum\limits_{j = 1}^N {\left\| {\varepsilon _j^{(v)}} \right\|_2^2} } \right)\\
Subject \; to: \begin{array}{*{20}{c}}
{}&{\left( {k \cdot {H^{(v)}}} \right) \cdot \beta }
\end{array} = t_i^T - {\left( {\varepsilon _i^{(v)}} \right)^T}\\
\begin{array}{*{20}{c}}
{}&{}&{\sum\limits_{v = 1}^V {{k^{r{\left( v \right)}}}}  = 1\begin{array}{*{20}{c}}
{}&{k > 0}
\end{array}}
\end{array}
\end{array}
\end{equation}
Iterative computing is used to solve both fusion coefficient and output weights due to generate interaction between feature fusion and optimization of hidden layer output weights. First, solve initial output weights by uniform fusing multi-view features. Second, testing samples of single feature with the help of solved weights. Finally, adjust the fusion coefficient according to errors of testing, moreover, repeating the firstly step. Therefore, we exploit Lagrange multiplier method, the equation is transformed into:
\begin{equation}
\begin{array}{l}
Minimize: \frac{1}{2}\left\| \beta  \right\|_2^2 + \sum\limits_{v = 1}^V {{k^{r{\left( v \right)}}}} \left( {\sum\limits_{j = 1}^N {\left\| {\varepsilon _j^{(v)}} \right\|_2^2} } \right)\\
Subject \; to: \begin{array}{*{20}{c}}
{}&{\left( {k \cdot {H^{(v)}}} \right) \cdot \beta }
\end{array} = t_i^T - {\left( {\varepsilon _i^{(v)}} \right)^T}\\
\begin{array}{*{20}{c}}
{}&{}&{\sum\limits_{v = 1}^V {{k^{r{\left( v \right)}}}}  = 1\begin{array}{*{20}{c}}
{}&{k > 0}
\end{array}}
\end{array}
\end{array}
\end{equation}

And then, ferivation both variables $\beta$, $k$ and Lagrange multipliers $a$, $b$, and obtain output weights display expression as follow:
\begin{equation}
\begin{array}{l}
Minimize: \frac{1}{2}\left\| \beta  \right\|_2^2 + \sum\limits_{v = 1}^V {{k^{r{\left( v \right)}}}} \left( {\sum\limits_{j = 1}^N {\left\| {\varepsilon _j^{(v)}} \right\|_2^2} } \right)\\
Subject \; to: \begin{array}{*{20}{c}}
{}&{\left( {k \cdot {H^{(v)}}} \right) \cdot \beta }
\end{array} = t_i^T - {\left( {\varepsilon _i^{(v)}} \right)^T}\\
\begin{array}{*{20}{c}}
{}&{}&{\sum\limits_{v = 1}^V {{k^{r{\left( v \right)}}}}  = 1\begin{array}{*{20}{c}}
{}&{k > 0}
\end{array}}
\end{array}
\end{array}
\end{equation}
, and the fusion coefficient:
\begin{equation}
\begin{array}{l}
Minimize: \frac{1}{2}\left\| \beta  \right\|_2^2 + \sum\limits_{v = 1}^V {{k^{r{\left( v \right)}}}} \left( {\sum\limits_{j = 1}^N {\left\| {\varepsilon _j^{(v)}} \right\|_2^2} } \right)\\
Subject \; to: \begin{array}{*{20}{c}}
{}&{\left( {k \cdot {H^{(v)}}} \right) \cdot \beta }
\end{array} = t_i^T - {\left( {\varepsilon _i^{(v)}} \right)^T}\\
\begin{array}{*{20}{c}}
{}&{}&{\sum\limits_{v = 1}^V {{k^{r{\left( v \right)}}}}  = 1\begin{array}{*{20}{c}}
{}&{k > 0}
\end{array}}
\end{array}
\end{array}
\end{equation}
In this paper, we improve above model to incremental detecting method. According to data generation from the cloud computing platform, the training process is divided into offline and online:
\par Offline training: sampling multi-view features from the cloud platform, training fusion model from accumulating data, and solve both $\beta$ and $k$ according to equation (9) and (10).
\par Online: the solved offline coefficient is used to fuse new dataset $x_{\Delta N}$, and obtain the updated output matrix. According to the equation (b), combine with the matrix $H_{\Delta N}$ to solve the incremental output weight matrix as follow:
\begin{equation}
\beta_{N+\Delta N} = S_1(H_NT_N+H_{\Delta N}^TT_{\Delta N}) = \beta_N+S_1H_{\Delta N}^T(T_{\Delta N}-H_{\Delta N}\beta_N)
\end{equation}
where $S_1^{-1}=S_0^{-1}+H_{\Delta N}^TH_{\Delta N}$. Therefore, our model achieves anomaly detecting in the cloud computing platform at real-time.
\subsection{Anomaly ranking and model adjusting}
Anomaly detecting is the precondition of anomaly handling to ensure high efficient running of the cloud computing platform. However, anomalies can be divided into difference degrees according to difference threats. Aiming to this problem, we proposed the novel method to ranking anomalies and the result of ranking is used to adjust the classification model to solve the imbalance problem. The workflow of the proposed model is shown in Fig. 2, including 4 steps: 1) training the classification model by the original training dataset, moreover, obtaining the output hidden weight and the output matrix; 2) according to statistical analysis, we find that it is closely related between the output matrix and the location of sample (verify it in the next paragraph). The classification matrix $T_{n\times c}$ can be obtained from the equation $H_{n\times m}\beta_{m\times c} = T_{n\times c}$, where the category number is the dimension of $T$, and the maximum value of any row of the matrix $f_max(T_{i\times c})$ is the class of the sample. We exploit the matrix $T$ to describe the location of each sample, achieving the sequence $l$ by ranking $f_max(T_{i\times c})$. The sequence $f_pos(f_max(T_{i\times c}))$ correspond the location sequence, which is smaller value corresponding the closer distance between the samples and the classification boundary; 3) weight samples ranked by $f_max(T_{i\times}c)/\sum_{i=1}^N{f_max(T_{i\times c})}$, due to the small number for abnormal samples, weighted samples can adjusting the adaptability of classification to imbalance distribution; 4) weighted samples are used to retrain the classification model to enhance the robustness.
\par In order to verify step 2, we obtain the statistics result of all of samples from the testing dataset. First, rank the vector $f_max(T_{i\times c})$ ; and then, divide the vector $f_max(T_{i\times c})$ into ten ranges according to values, and count the number for each range. The more intuitive is shown in Fig. 3, the right of figure 3 is the histogram that is the samples number from then evenly-distributed ranges, and the left of figure 3 is the scatter-plots that is corresponding the histogram. From the figure 3, we can know locations from three kinds of samples, in 1th type, the number of 20\% samples in front of the vector $l_head$ is rather less and samples close to the classification boundary. In 2th type, the number of 20\% samples in rear of the vector $l_rear$ is rather less and samples far from the classification boundary. In 3th type, remaining samples that account for the largest proportions and these samples locate in the medium position of the classification region. Therefore, above conclusion that is the vector $f_max(T_{i\times c})$ representing the location of the sample is verified by analysis.
\begin{figure}
  \includegraphics[width = 11cm]{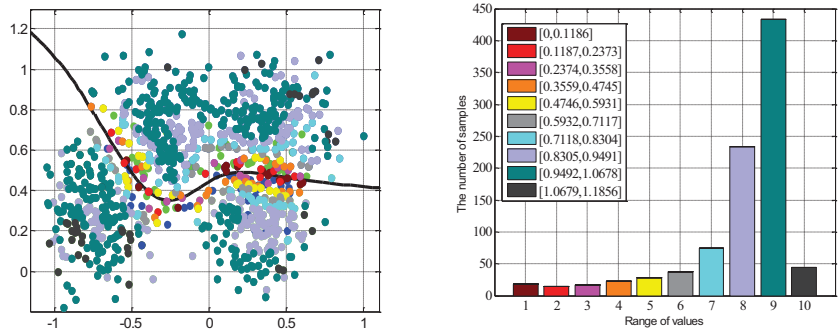}
\caption{Statistical analysis of samples distribution}
\label{fig:1}       
\end{figure}
\section{Experiments}
\subsection{Experiments setting}
In order to measure the performance of the proposed anomaly detecting method, we deploy the experiment environment based on OpenStack, the framework is shown in Fig.4. Moreover, the specific is set as the following:
(1)	The operating system is Ubuntu16.4.01 in the virtual service of the master and services, and we install the virtualization hypervisor KVM+Qemu.
(2)	The manage component is OpenStack in the cloud computing platform.
(3)	Performance indicator vectors of virtual services are collected from Prometheus.
(4)	In order to simulate the working state of each virtual service, Sysbench and Webbench are used benchmarking frameworks.
We choose 5 computers as work nodes, and deploy 3 virtual services in each node.
\subsection{Dataset collecting and descripting}
In this paper, we grab the state data of the virtual service from Metric Agent service to Retrieval in real-time. In order to avoid network congestion and to ensure transmission efficiency of computing data, we collect samples under 10s frequency, and sustained collection 3 hours. Therefore, we obtain the total items is 16200 where offline samples 20\% of all samples (3240 items). Online detecting sustains 36 minutes where the number of training samples is 1620 and the number of testing samples is 1620.
We collect 4 kinds of state data from virtual services including CPU, memory, disk I/O, network-service. Each item is described as follows <time-stamp, host, attribute-set (CPU), attribute-set (memory), attribute-set (disk-I/O), attribute-set (network-service). The specific description of attributes is shown in table 1.
\newcommand{\tabincell}[2]{\begin{tabular}{@{}#1@{}}#2\end{tabular}}
\begin{table}
\caption{The attributes description (CPU and Memory) of sub-systems}
\label{tab:1}       
\begin{tabular}{llll}
\hline\noalign{\smallskip}
\tabincell{c} {CPU \\attributes set} & Description & \tabincell{c}{Memory \\attributes set} & Description\\
\noalign{\smallskip}\hline\noalign{\smallskip}
\tabincell{l}{node\_\\cpu\_idle} & Free percentage & \tabincell{l}{node\_\\memory} & Available memory\\
\tabincell{l}{node\_\\cpu\_iowait} & Wating I/O time & \tabincell{l}{node\_\\memory\_Buffers} & Block device cache \\
\tabincell{l}{node\_\\cpu\_softing} & Response software interruption time & \tabincell{l}{node\_\\memory\_Cached} & Character device cache\\
\tabincell{l}{node\_\\cpu\_system} & Proportion of Kernel Operations & \tabincell{l}{node\_\\memory\_Swapd} & Number of Use Spaces\\
\tabincell{l}{node\_\\cpu\_user} & User Process Propo & \tabincell{l}{node\_\\memory\_MemTotal} & Total physical memory\\
\tabincell{l}{node\_\\cpu\_nice} & Change process & \tabincell{l}{node\_\\memory\_Memfree} & Idle number\\
\tabincell{l}{node\_\\cpu\_irq} & Response to hardware interrupt time & \tabincell{l}{node\_\\memory\_Slab} & Kernel uses memory\\
\tabincell{l}{node\_\\cpu\_cs} & Process switching time & \tabincell{l}{node\_\\memory\_Sheme} & Process shared memory \\
\tabincell{l}{node\_\\cpu\_running} & Number of Runnable Tasks & \tabincell{l}{node\_\\memory\_VmallocTotal} & Virtual Machine Memory Volume\\
\tabincell{l}{node\_\\vcpu\_run} & Virtual Machine Runtime & \tabincell{l}{node\_\\memory\_VmallocRate} & Virtual Machine Memory Utilization Rate \\
\tabincell{l}{node\_\\cpu\_runrate} & Virtual Machine Utilization Rate & \tabincell{l}{node\_\\memory\_VmallocMax} & AMaximum occupancy of virtual machines\\
\noalign{\smallskip}\hline
\end{tabular}
\end{table}

\begin{center}
\begin{table}
\caption{The attributes (I/O and Network) description of sub-systems}
\label{tab:1}       
\begin{tabular}{llll}
\hline\noalign{\smallskip}
\tabincell{c} {I/O \\attributes set} & Description & \tabincell{c}{Network \\attributes set} & Description\\
\noalign{\smallskip}\hline\noalign{\smallskip}
\tabincell{l}{node\_\\disk\_await} & \tabincell{l}{I/O \\waiting time} & \tabincell{l}{node\_\\network}& \tabincell{l}{The amount of\\ data received per second}\\
\tabincell{l}{node\_\\disk\_svc\_time} & \tabincell{l}{I/O \\service time} & \tabincell{l}{node\_\\network\_transmit\_bytes} & \tabincell{l}{The amount of \\data sended per second}\\
\tabincell{l}{node\_\\disk\_read\_time\_ms} & \tabincell{l}{Number of \\readings per second} & \tabincell{l}{node\_\\network\_receive\_packets} & \tabincell{l}{Packages received \\per second}\\ \tabincell{l}{node\_\\disk\_write\_time\_ms} & \tabincell{l}{Number of \\writing per second} & \tabincell{l}{node\_\\network\_transmit\_packets} & \tabincell{l}{The amount of \\ data sended per second}\\
\tabincell{l}{node\_\\disk\_sectors\_written} & \tabincell{l}{Reading \\sector count} & \tabincell{l}{node\_\\network\_trLoss\_packets} & \tabincell{l}{Number of \\Packets Lost on Acceptance}\\
\tabincell{l}{node\_\\disk\_sectore\_written} & \tabincell{l}{Writing \\sector count} & \tabincell{l}{node\_\\network\_trLoss\_packets} &\tabincell{l}{Number of \\ Packets Lost When Sending}\\ \tabincell{l}{node\_\\disk\_io\_time\_weighted} & \tabincell{l}{Percentage of \\ operating time} & \tabincell{l}{node\_\\netstat\_TcpExt\_TCPOFOQueue} & \tabincell{l}{TCP\\ sequence}\\
\tabincell{l}{node\_\\disk\_bytes\_read} & \tabincell{l}{Reading\\ bit number} & \tabincell{l}{node\_\\netstat\_TcpExt\_TCPOrigDataSent} & \tabincell{l}{TCP \\traffic}\\ \tabincell{l}{node\_\\disk\_bytes\_written} & \tabincell{l}{Writing \\bit number} & \tabincell{l}{node\_\\netstat\_TcpExt\_TCPLos} & \tabincell{l}{TCP \\untraffic}\\
\tabincell{l}{node\_\\disk\_Vread\_time} & \tabincell{l}{Number of \\virtual block reads per second} & \tabincell{l}{node\_\\Vnetwork\_receive\_bytes} & \tabincell{l}{Virtual Network \\ Accepts Data Volume}\\
\tabincell{l}{node\_\\disk\_Vwrite\_time} & \tabincell{l}{Write times \\per second for virtual blocks} & \tabincell{l}{node\_\\Vnetwork\_transmit\_bytes} & \tabincell{l}{The amount of\\ data sent by virtual network}\\
\noalign{\smallskip}\hline
\end{tabular}
\end{table}
\end{center}
CPU anomaly: run computation programs in virtual services to achieve the very high CPU utilization rate. In this paper, the CPU anomaly is repressed as cpu\_Calculation.
I/O anomaly: creating, writing, reading a large number of files in virtual services achieve system I/O anomalies. In this paper, the I/O anomaly is repressed as io\_Operate.
Network anomaly: send a large number of requiring to achieve network anomalies injection and induce the high occupancy rate of network resources of virtual services. In this paper, the Network anomaly is repressed as net\_Operate.
Memory anomaly: reading/writing the fixed-size memory block to increase the memory load. In this paper, memory anomaly include two classes: memory\_Read and memory\_Write.

\subsection{Experiment results and analysis}
Fig.2. is the histogram of sample size, and compares the positive and negative of dataset due to datasets contained anomalies are imbalance distribution.
\begin{figure}
  \includegraphics[width = 8cm]{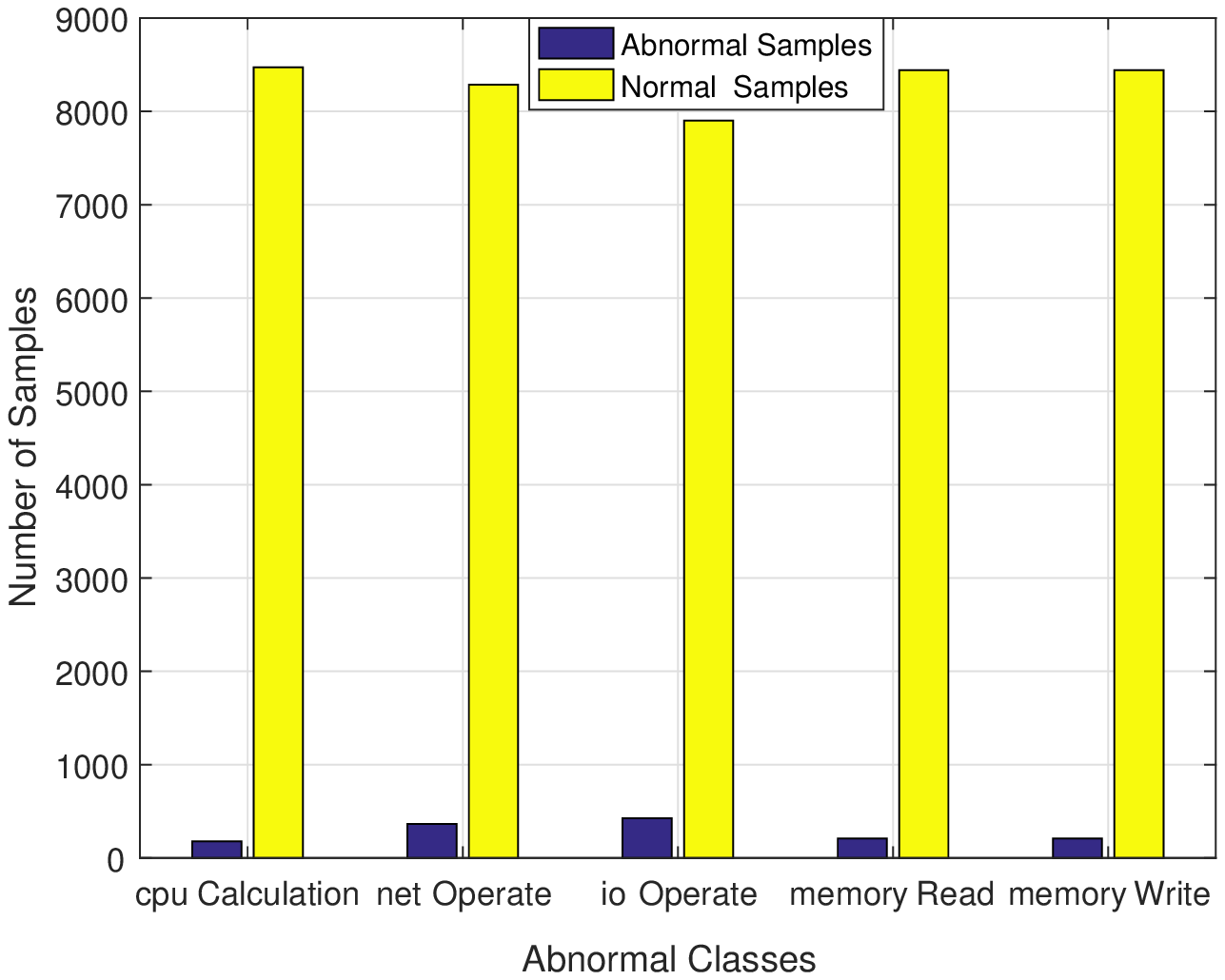}
\caption{Please write your figure caption here}
\label{fig:1}       
\end{figure}
In order to verify the performance of anomaly detecting in cloud computing platform, we utilize ROC curve to visual represent experiment results, therefore, we obtain both (false positive rate, RateFP) and (true positive rate, RateTP) of each sample, the defined as follows:
\begin{equation}
Rate_TP = E_TP/E_P; Rate_FP=E_FP/E_N
\end{equation}
where the number of positive is $E_P$, and the number of negative is $E_N$. $E_TP$ and $E_TN$ that are the number of classification results are used to compose the coordinate of ROC curve. ROC curve describe the performance by plotting points and linking these points as a curve. When the location of the point is closer to 1, we can know that the classification achieves more performance.
In this paper, we choose 3 models to compare and analysis the effectiveness of the proposed model. (1) OrigiF-KNN model, link all features from all subsystems from the cloud computing platform directly, and then KNN model is used to detect anomalies. (2) PCA-KNN model, utilize PCA to reduce the dimension of the linked feature, and then KNN model is used as the classification model. (3) KernelPCA-KNN model, utilize kernel PCA to reduce the dimension of the linked feature, and then KNN model is used as the classification model. Fig.6. shows experiment results including our model and above 3 models, where blue curves are detection results from normal samples, and yellow curves are detection results from anomalies.
From figure 7, the proposed method enhance the detecting accuracy due to fully utilizes the complementary information from difference subsystems. Figure 8 is comparison between difference models when cpu\_calculation is injected in the cloud computing platform. We can learn that Origi-KNN obtains the unsatisfied result due to the linked feature contains a lot of noisy and redundancy information to distribute the detecting result. KernelPAC-KNN by non-linear mapping obtains higher performance comparing with PCA-KNN. Fig. 9. shows difference detecting results under io\_Operate anomalies, our model combines with the supervised information to achieve multi-view features automatic fusion, and optimize the solution space of the classification model to enhance the performance. When net\_Operate anomalies are injected in the cloud computing platform, curves of Origi-KNN and KernelPCA-KNN are closer to the diagonal position of ROC, then detecting accuracies of both models are random from figure 9. The proposed model obtain the satisfied performance. In figure 10, we inject both memory\_Read and memory\_Write in the cloud computing platform. PCA-KNN, KernelPCA-KNN and our model achieve better results under the kind of anomaly.

\begin{figure}[!t]
\centerline{
\subfloat[]{\includegraphics[width=2in]{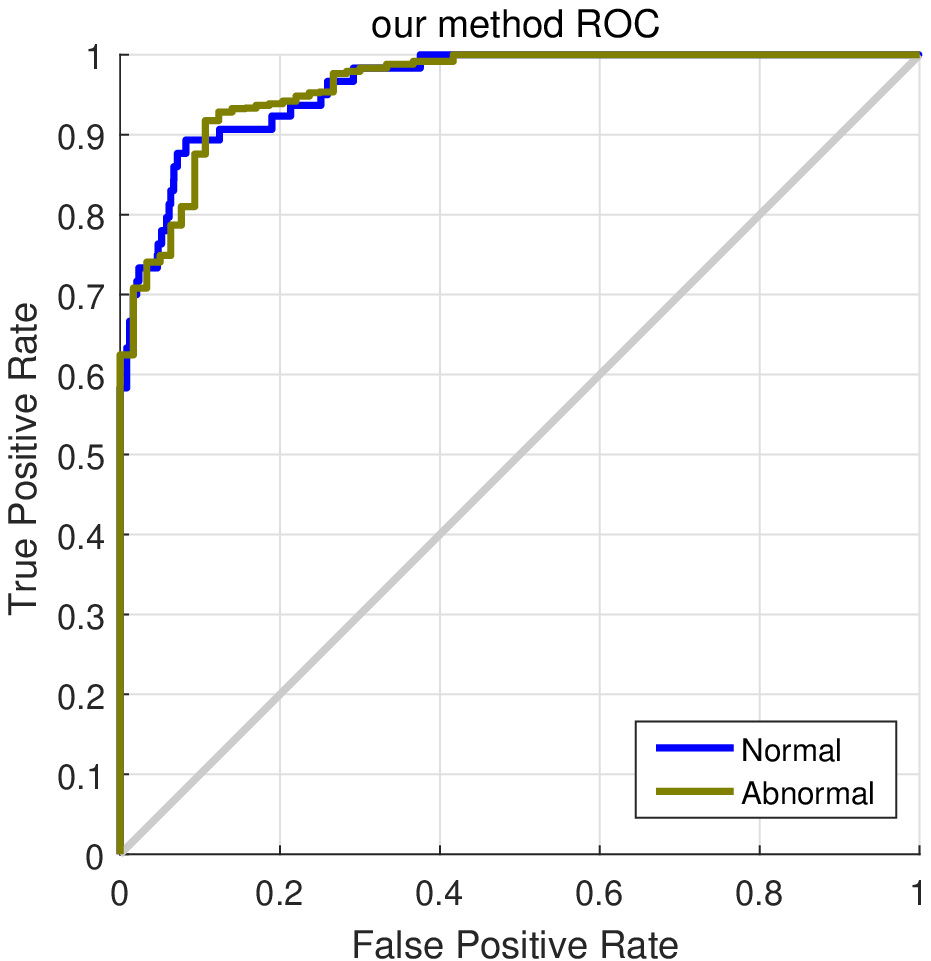}}
\subfloat[]{\includegraphics[width=2in]{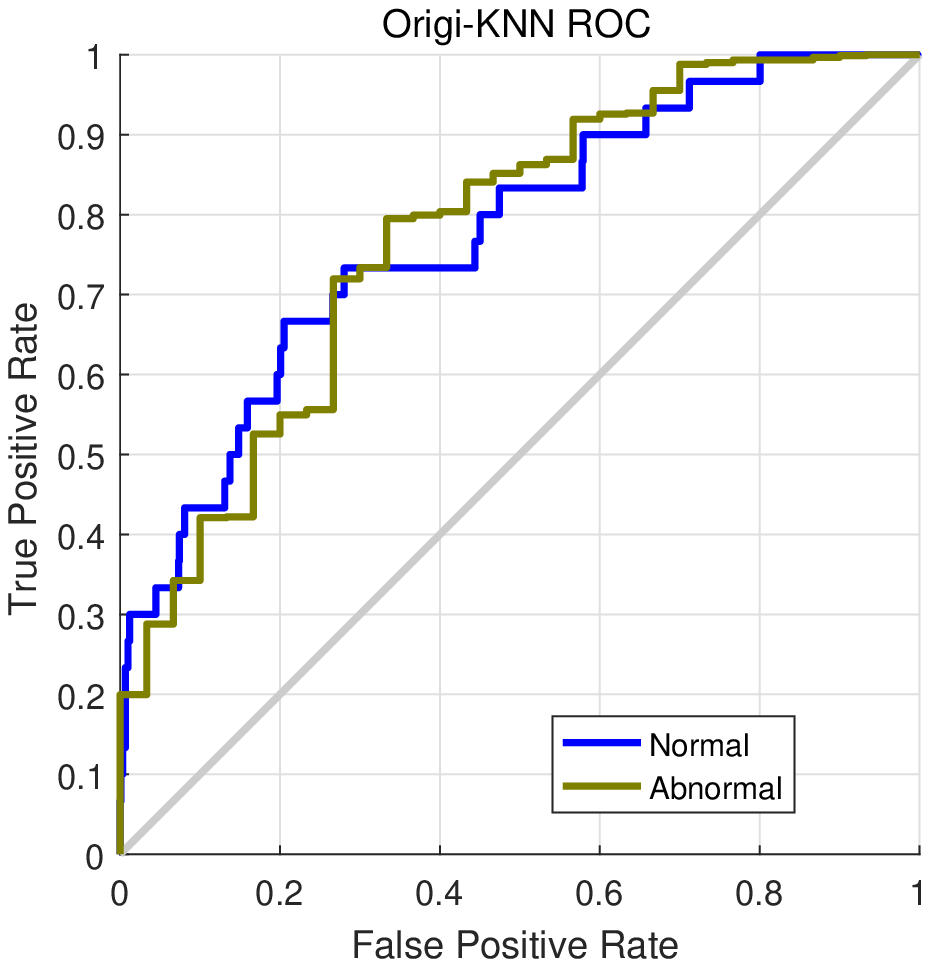}}}
\centerline{
\subfloat[]{\includegraphics[width=2in]{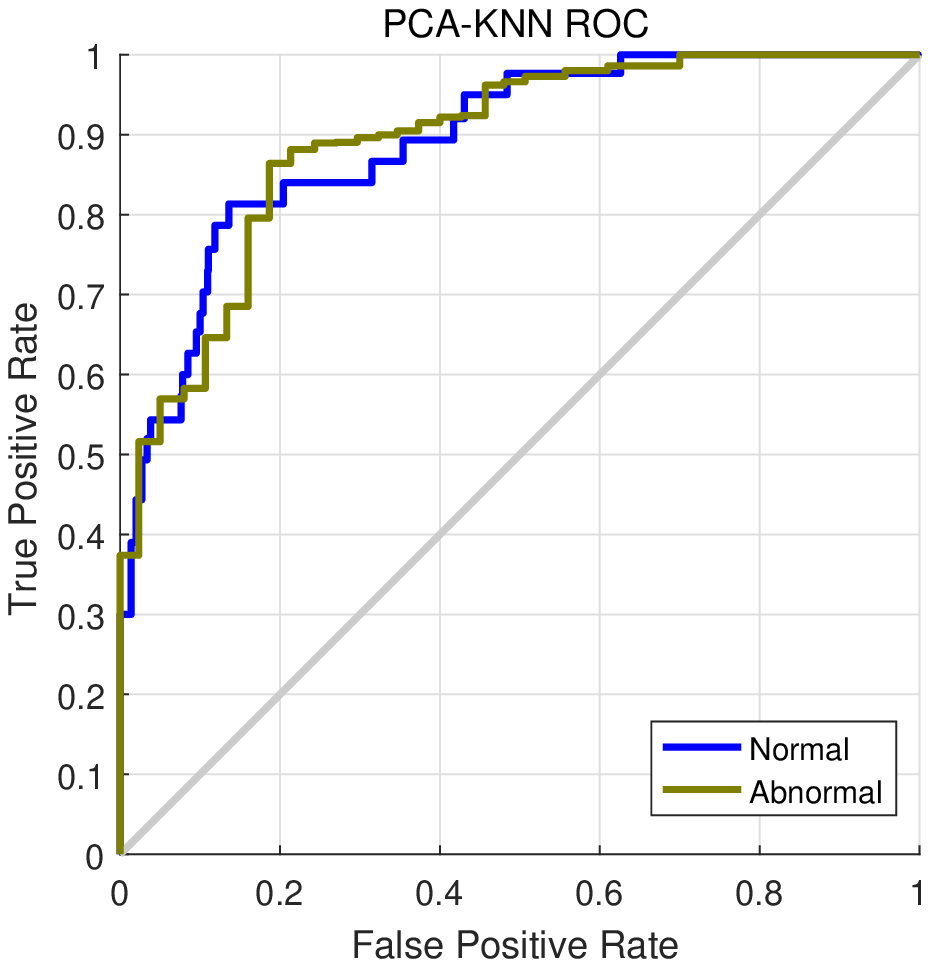}}
\subfloat[]{\includegraphics[width=2in]{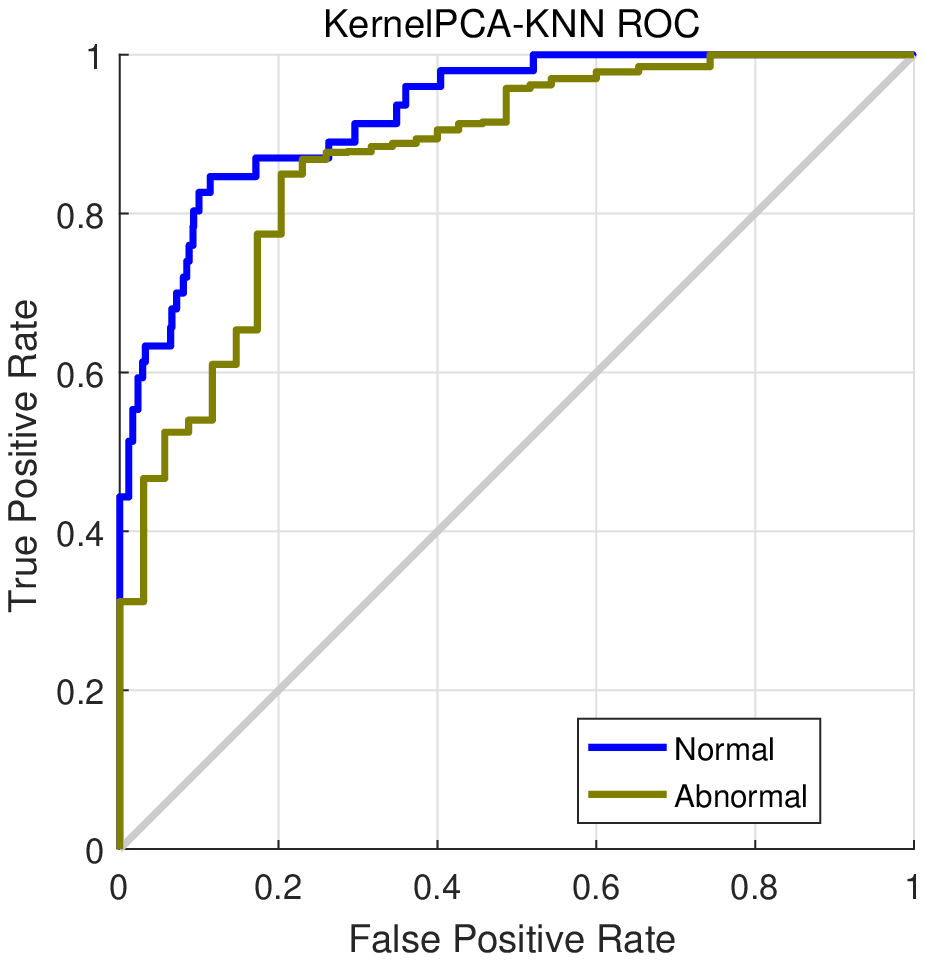}}}
 \caption{Comparison of ROC curves of cpu\_Calculation anomalies}
 \label{fig_sim}
\end{figure}

\begin{figure}[!t]
\centerline{
\subfloat[ ]{\includegraphics[width=2in]{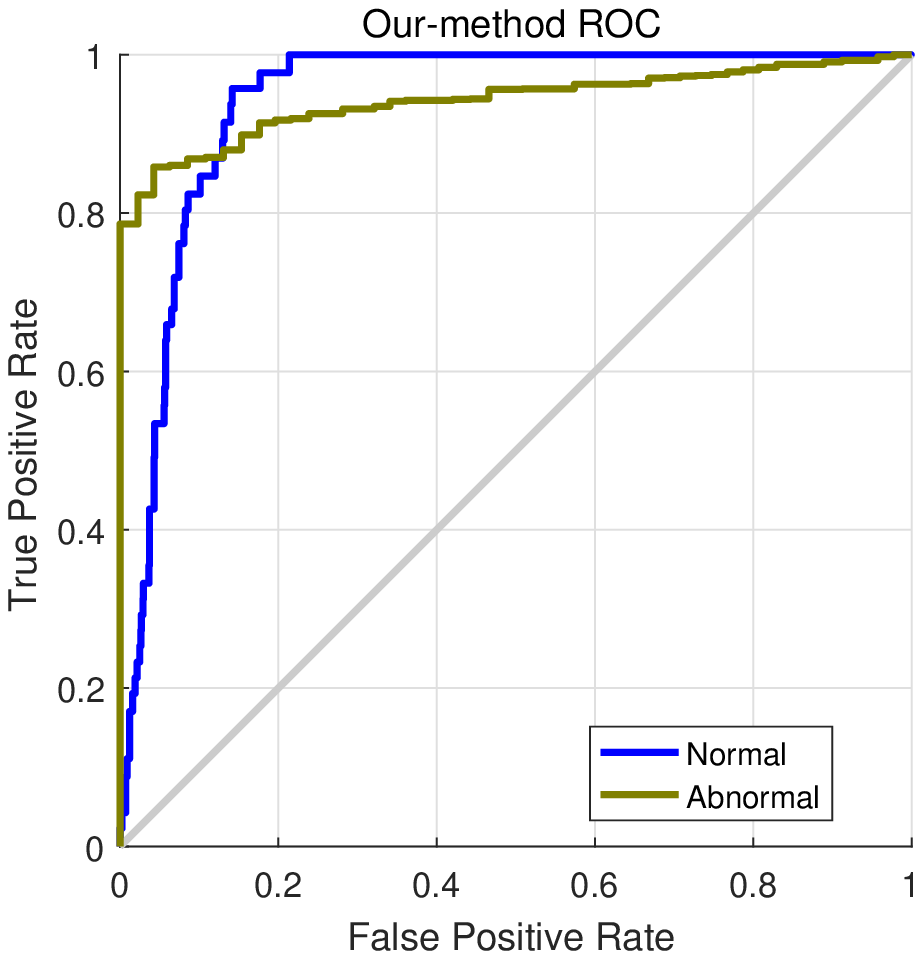}}
\subfloat[ ]{\includegraphics[width=2in]{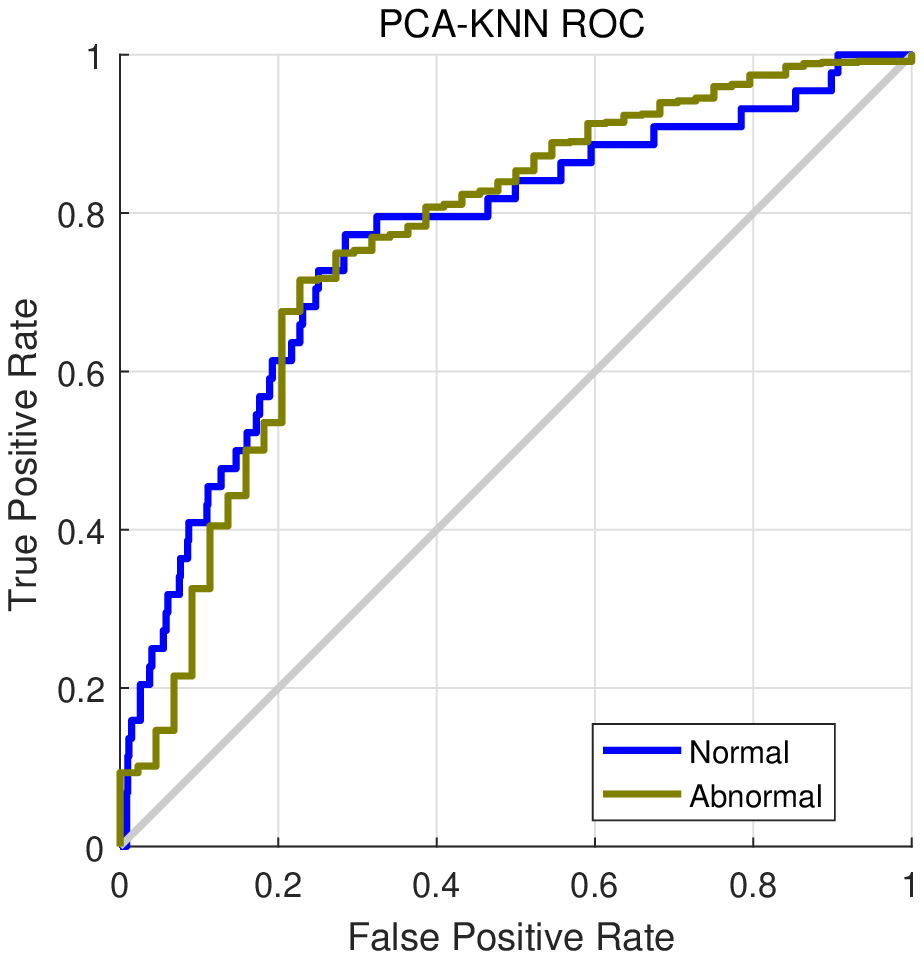}}}
\centerline{
\subfloat[ ]{\includegraphics[width=2in]{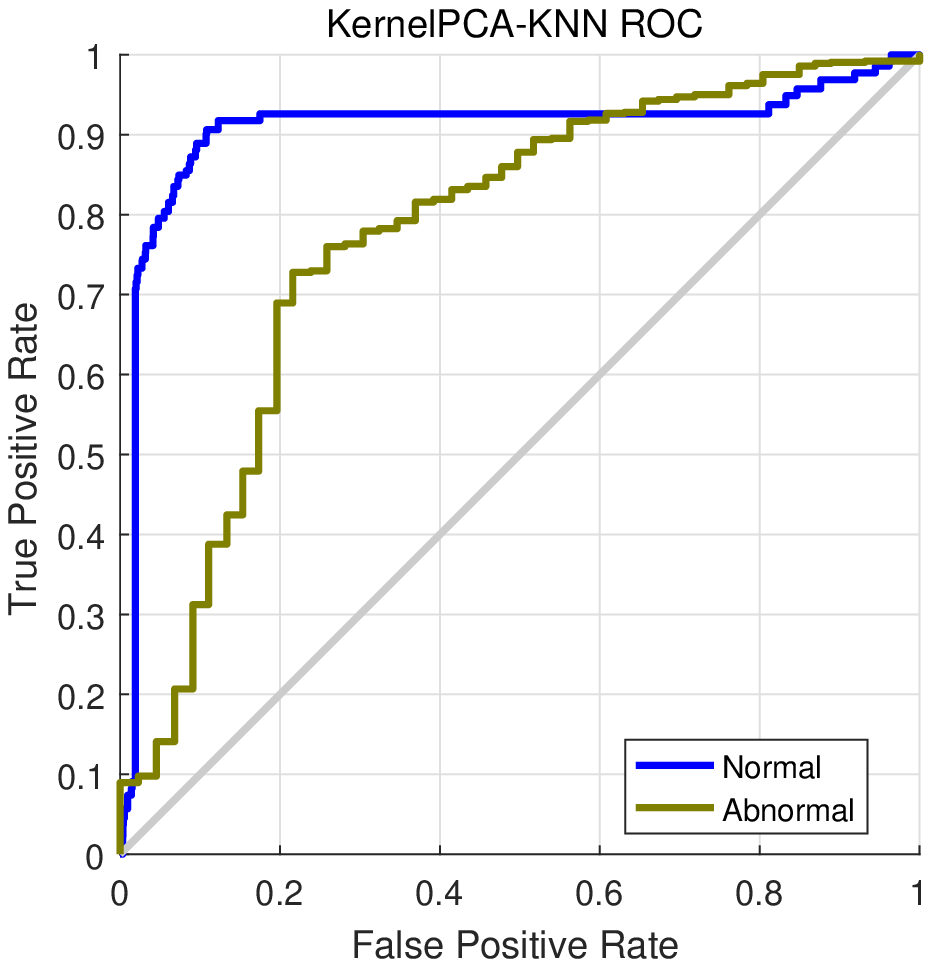}}
\subfloat[ ]{\includegraphics[width=2in]{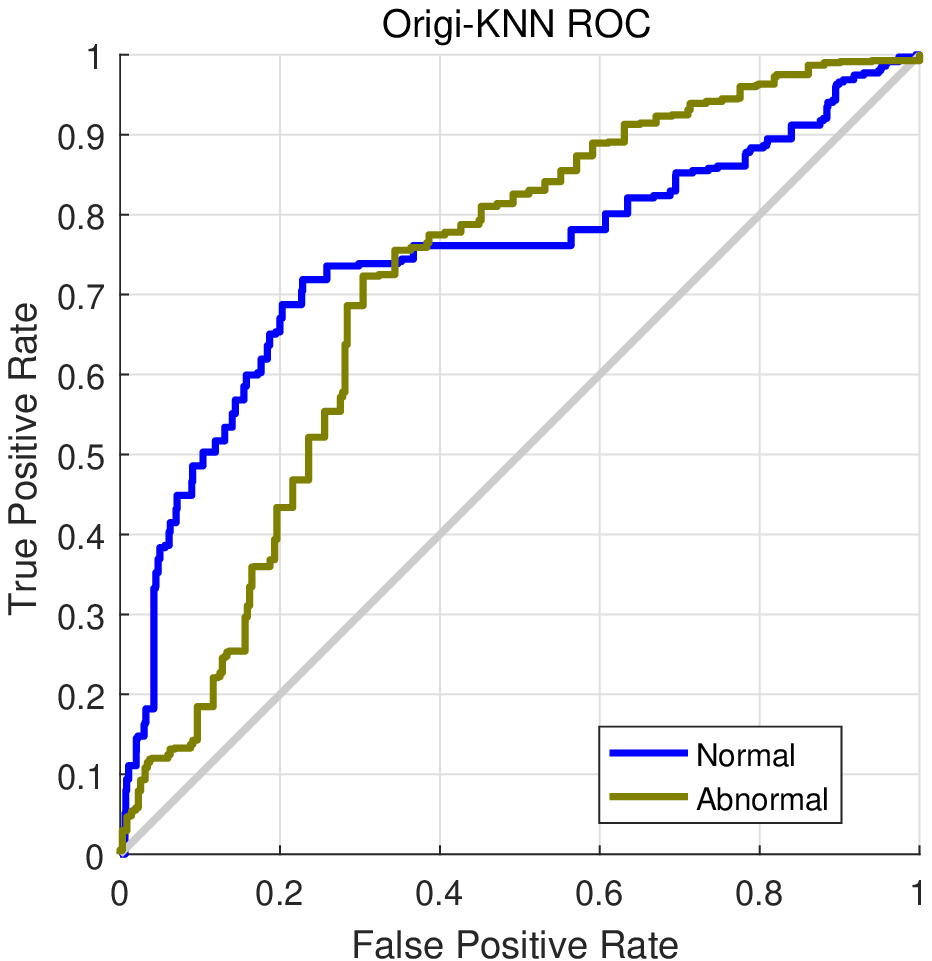}}}
 \caption{Comparison of ROC curves of io\_Operate anomalies}
 \label{fig_sim}
\end{figure}

\begin{figure}[!t]
\centerline{
\subfloat[ ]{\includegraphics[width=2in]{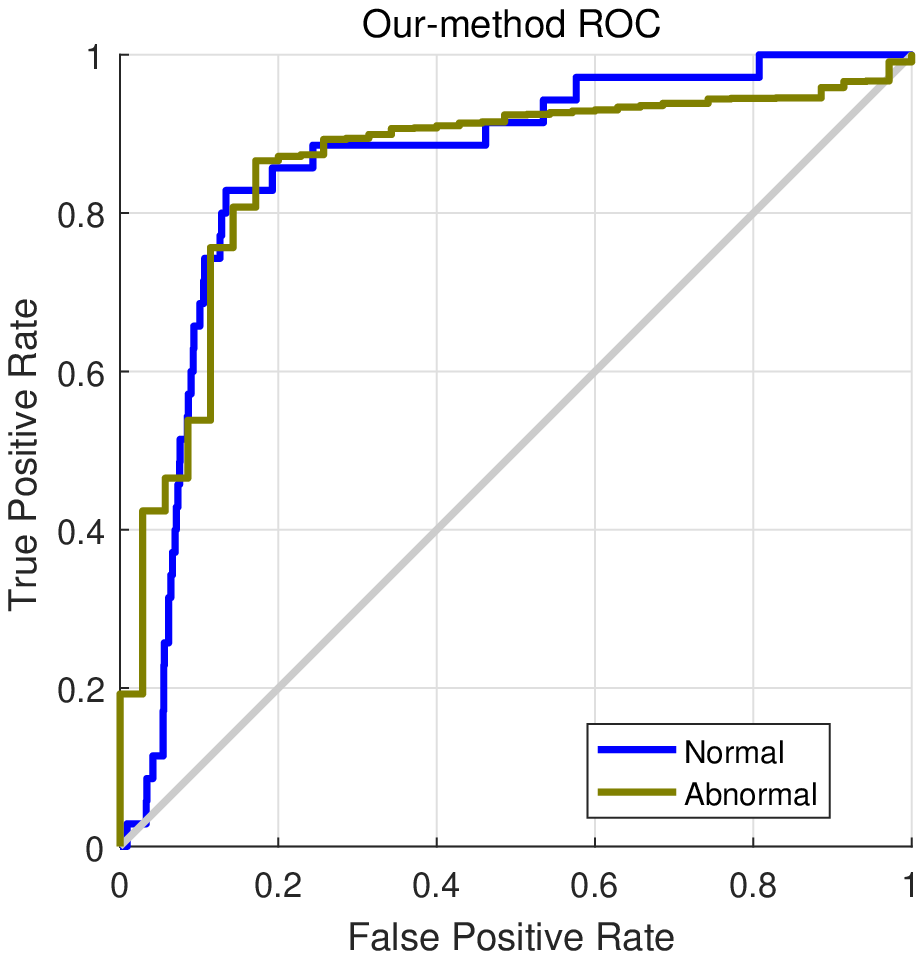}}
\subfloat[ ]{\includegraphics[width=2in]{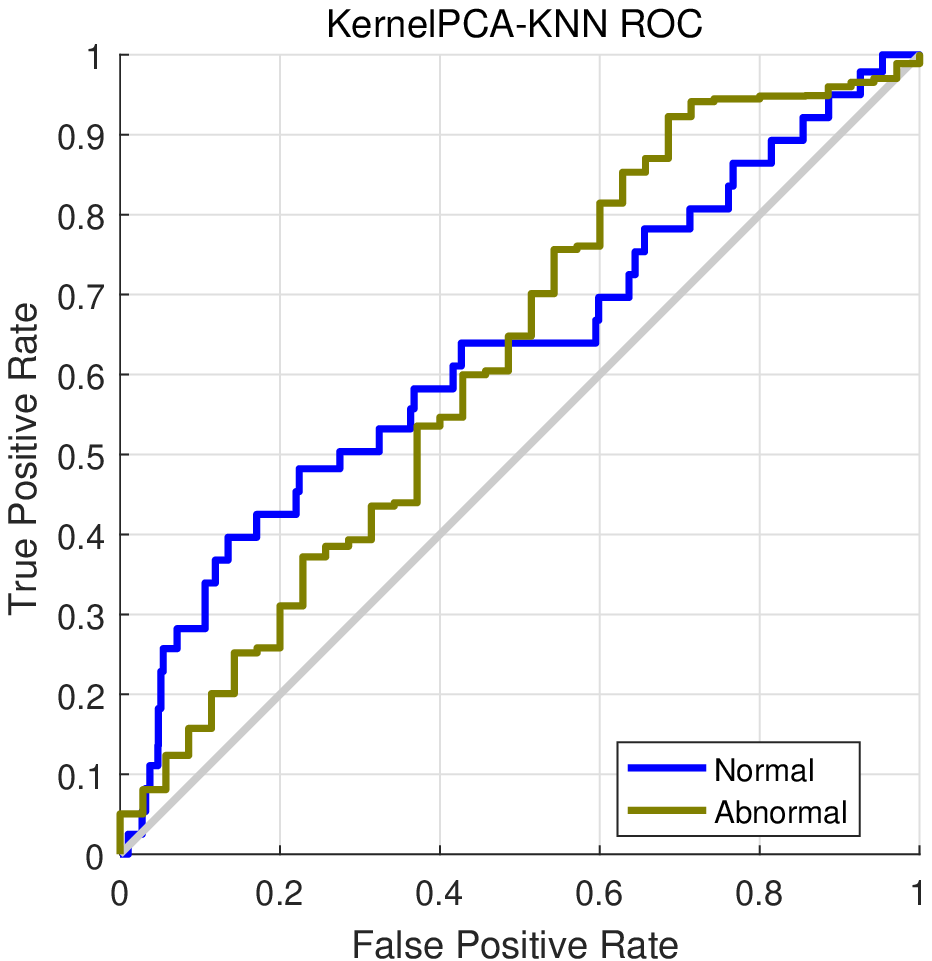}}}
\centerline{
\subfloat[ ]{\includegraphics[width=2in]{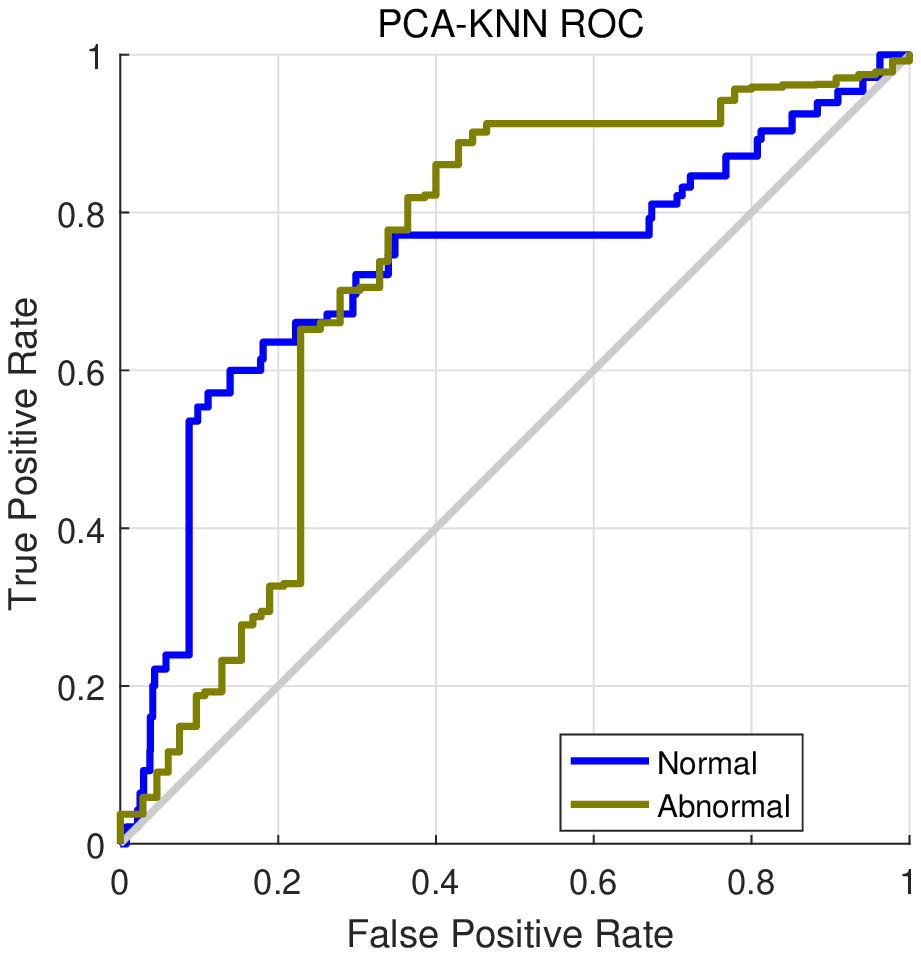}}
\subfloat[ ]{\includegraphics[width=2in]{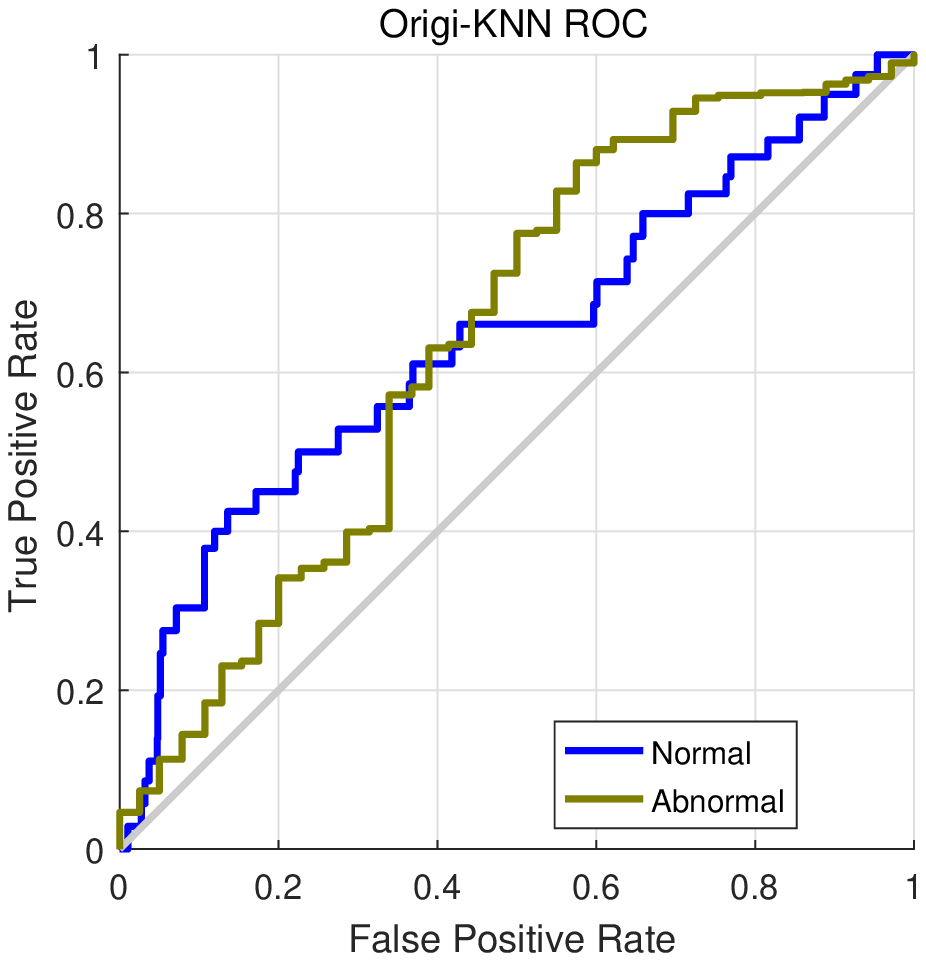}}}
 \caption{Comparison of ROC curves of net\_Operate anomalies}
 \label{fig_sim}
\end{figure}

\begin{figure}[!t]
\centerline{
\subfloat[ ]{\includegraphics[width=2in]{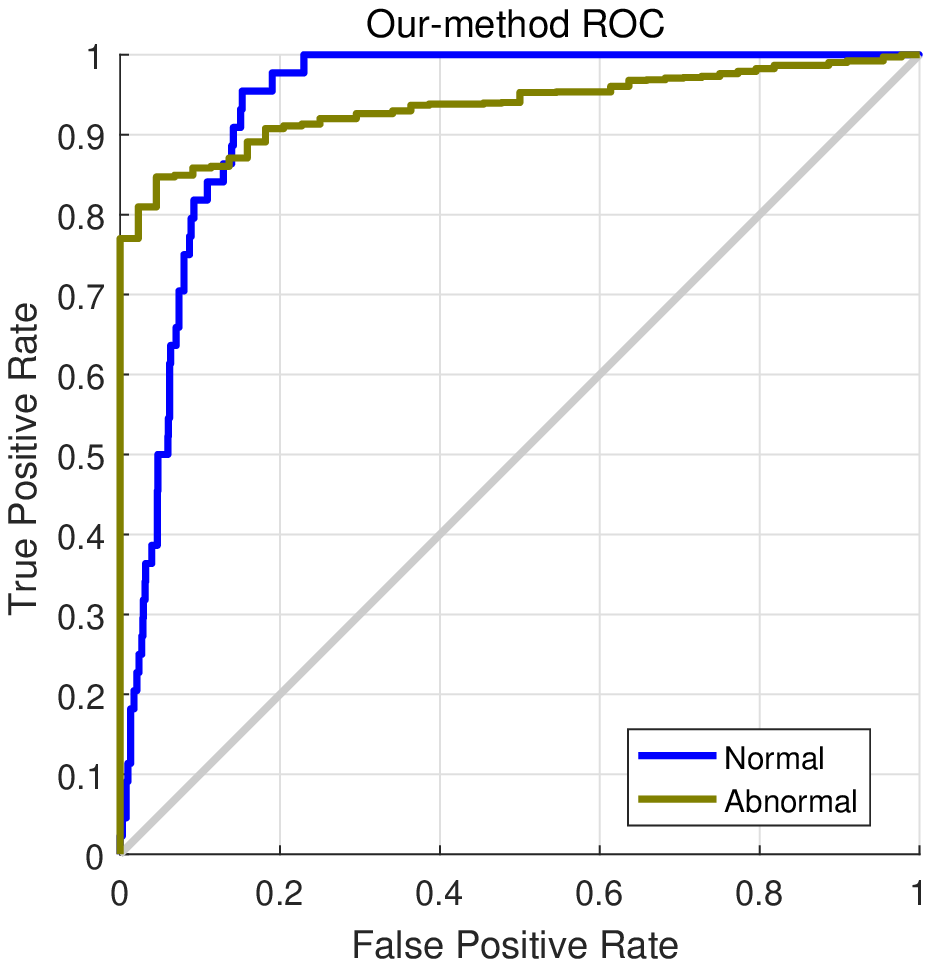}}
\subfloat[ ]{\includegraphics[width=2in]{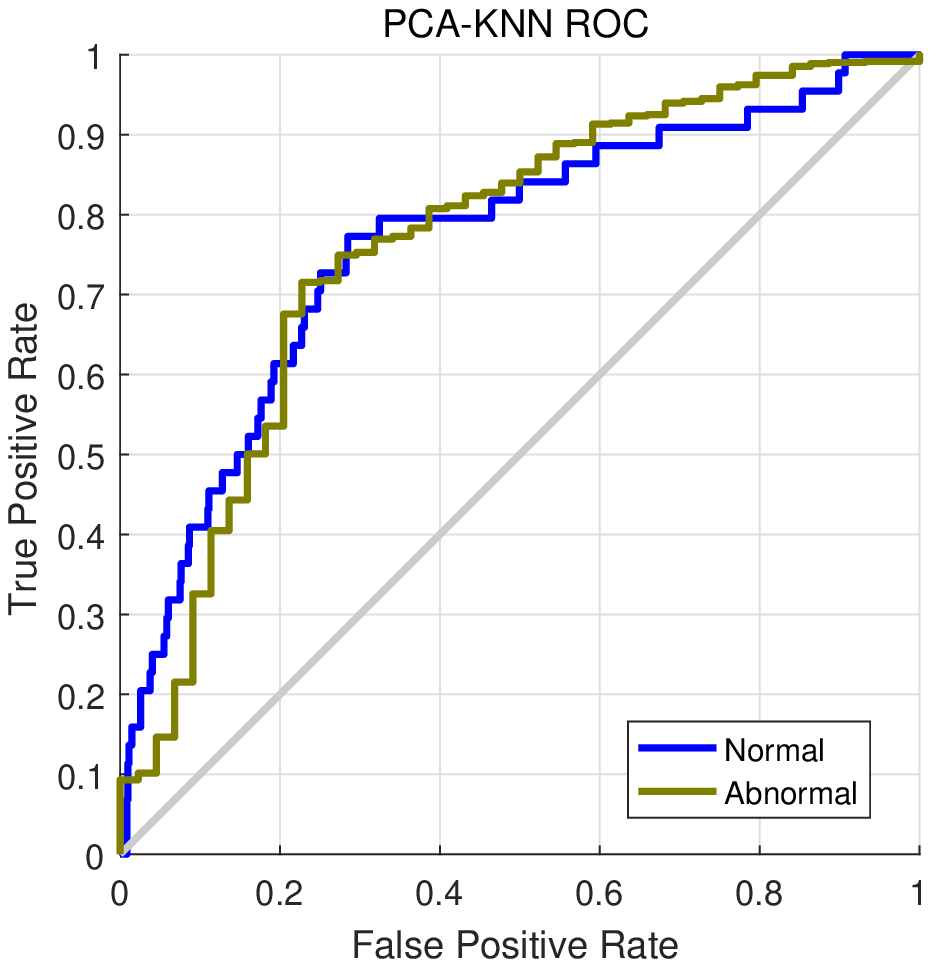}}}
\centerline{
\subfloat[ ]{\includegraphics[width=2in]{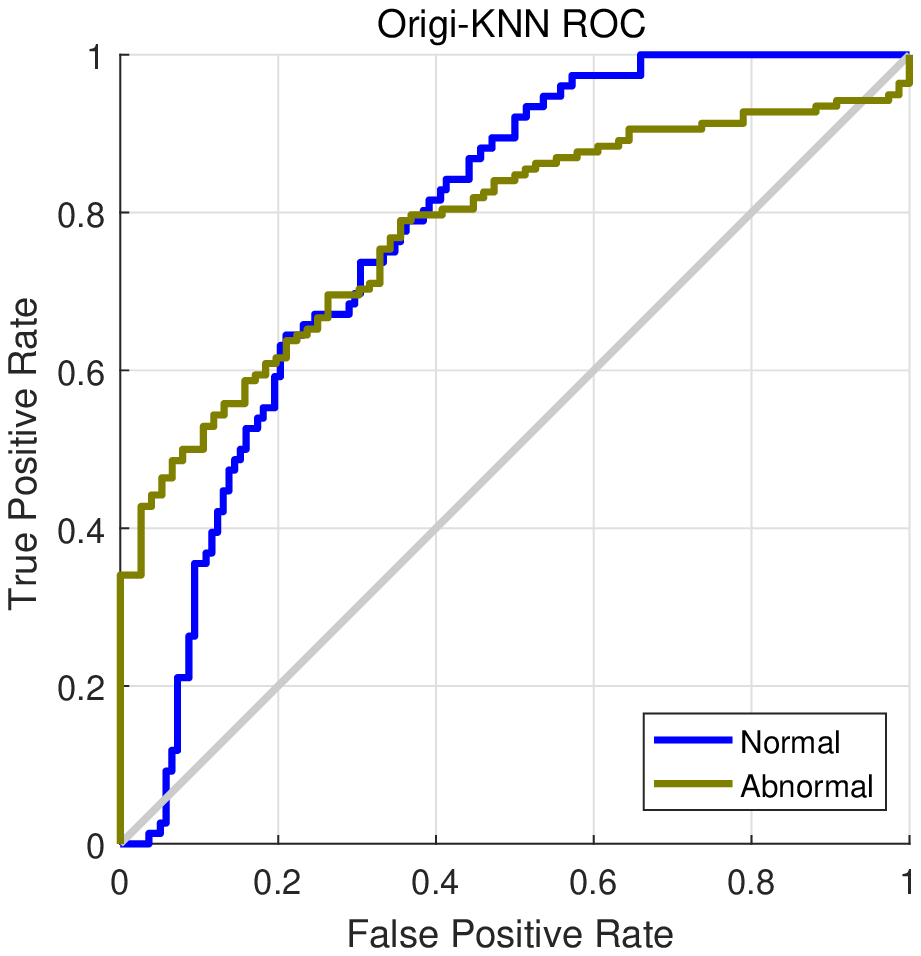}}
\subfloat[ ]{\includegraphics[width=2in]{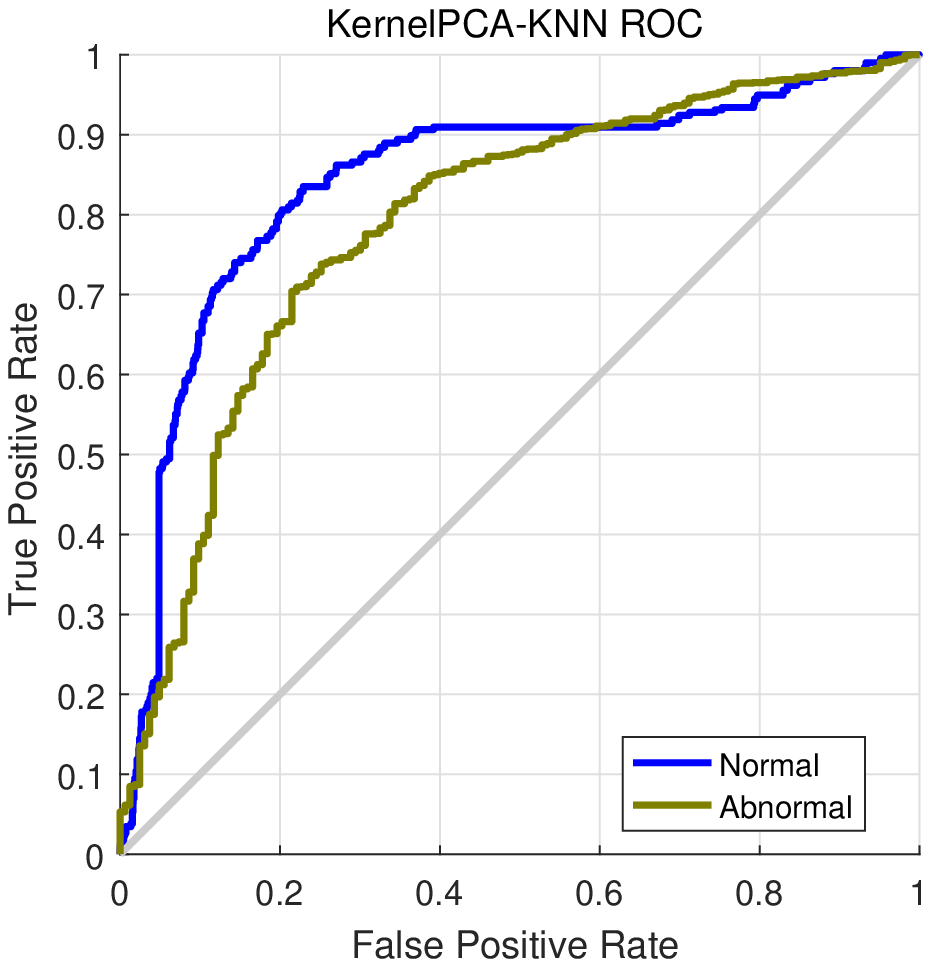}}}
 \caption{Comparison of ROC curves of memory\_Read anomalies}
 \label{fig_sim}
\end{figure}

\begin{figure}[!t]
\centerline{
\subfloat[ ]{\includegraphics[width=2in]{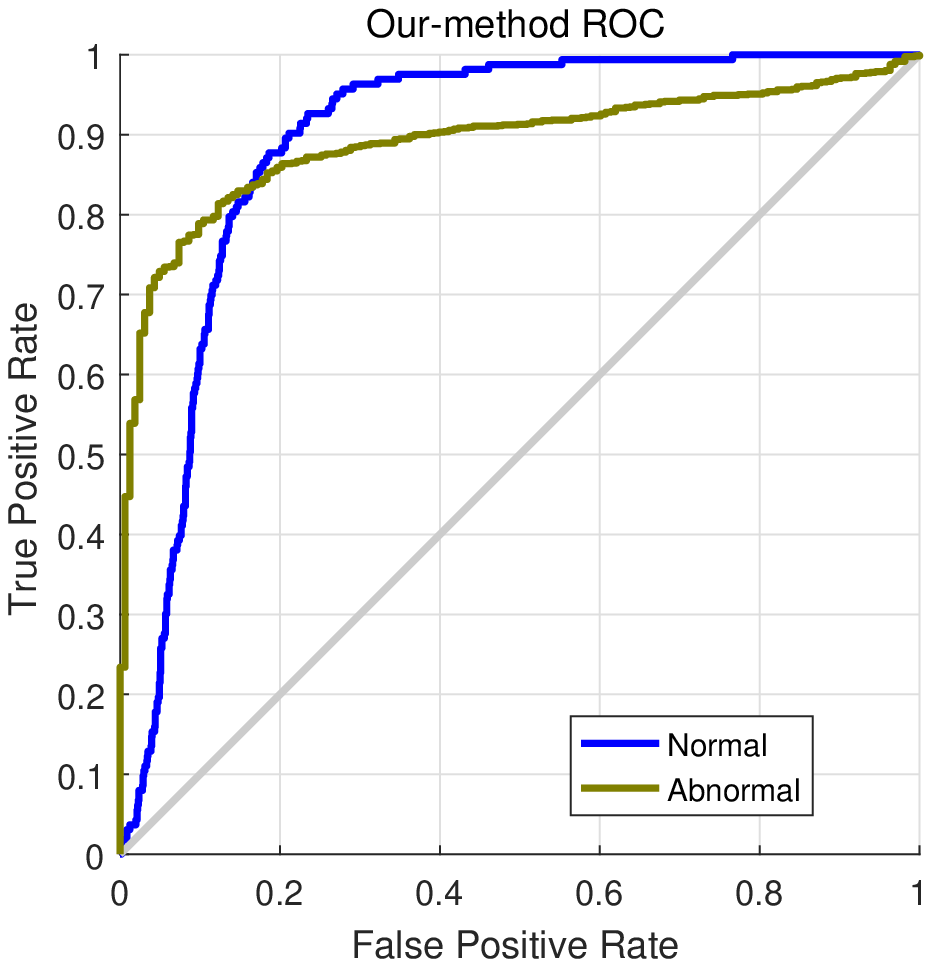}}
\subfloat[ ]{\includegraphics[width=2in]{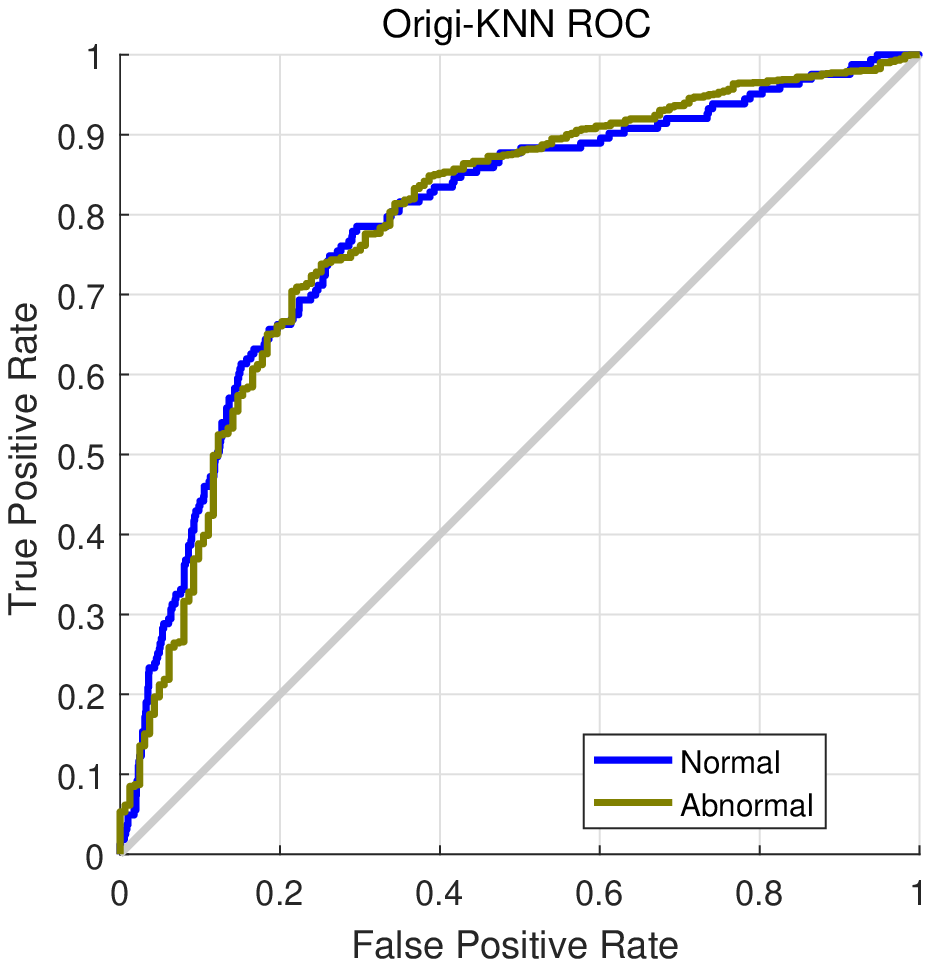}}}
\centerline{
\subfloat[ ]{\includegraphics[width=2in]{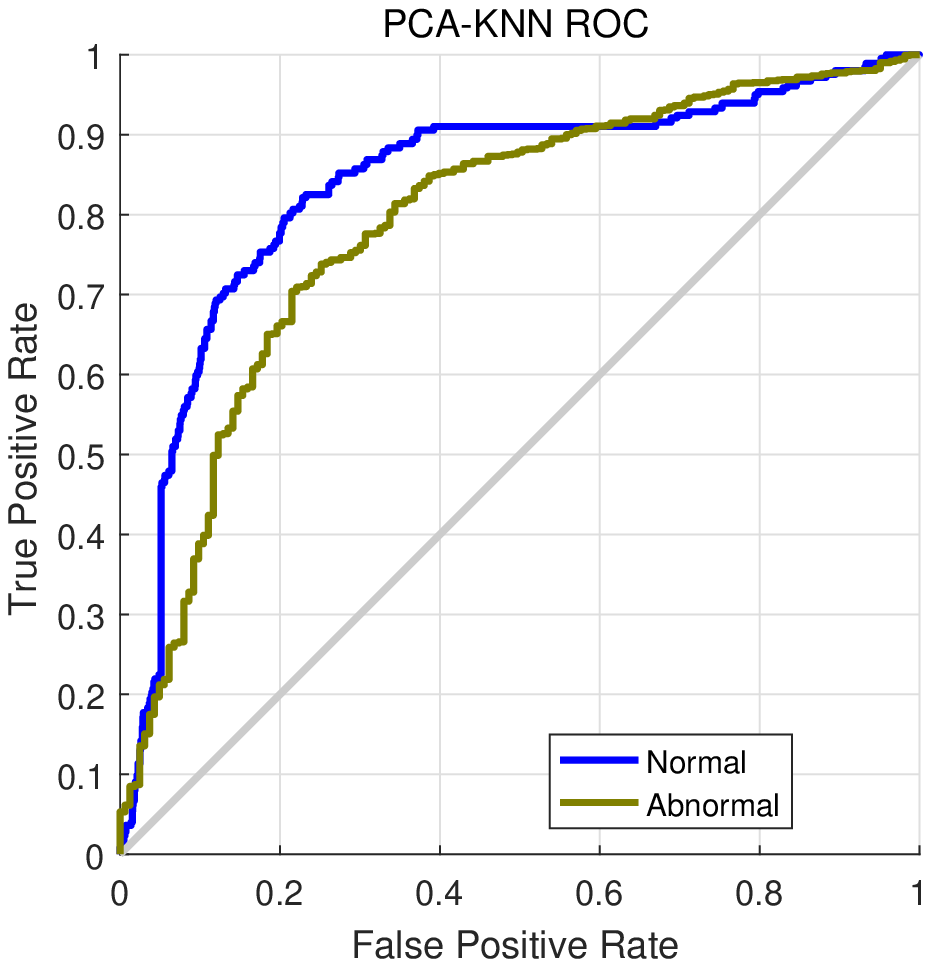}}
\subfloat[ ]{\includegraphics[width=2in]{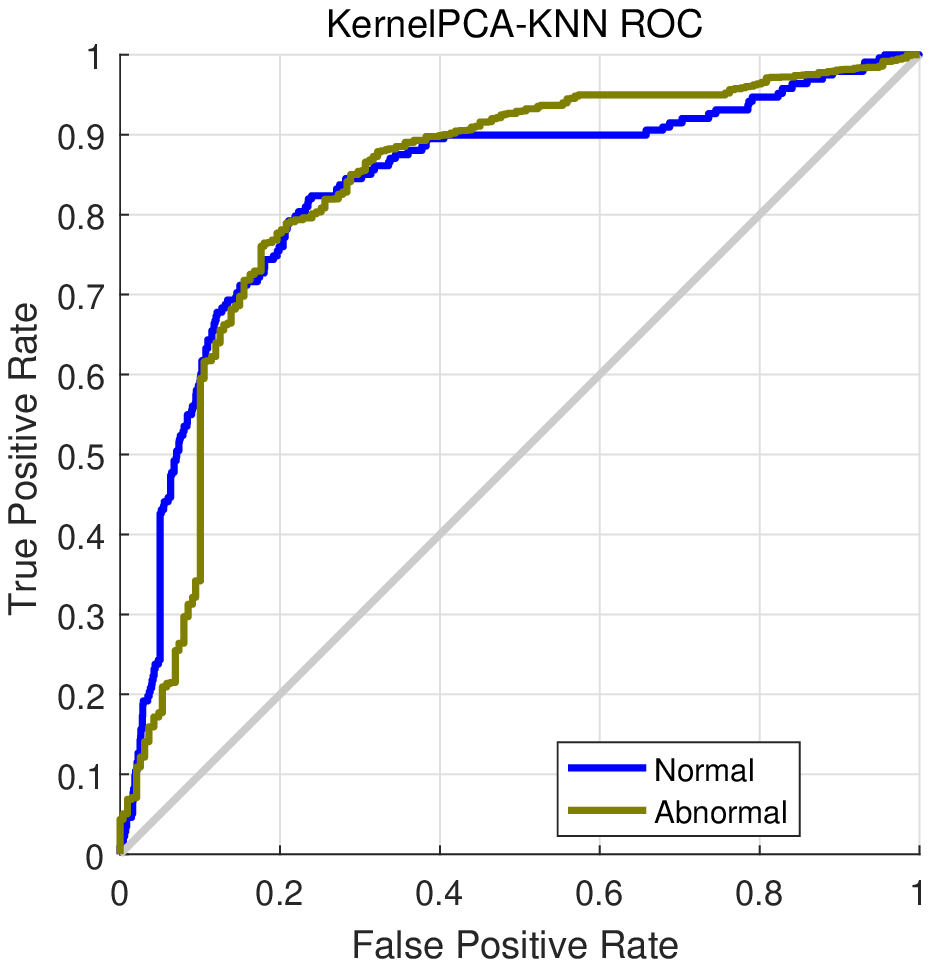}}}
 \caption{Comparison of ROC curves of memory\_Write anomalies}
 \label{fig_sim}
\end{figure}

\section{Conclusion}
We presented an effective method for anomaly detecting of the cloud computing platform by building the incremental machine learning model. Our problem is formulated as the binary classification problem in real-time, whose solution is learned through a improve multiple features ELM model. The proposed model automatic fuses multiple features from difference sub-systems and obtains the optimized classification solution by minimizing the training error sum; ranked anomalies are determined by the relation between samples and the classification boundary, and weighting samples ranked retrain the classification model. We can deal with various challenges in anomaly detecting, such as imbalance distribution, high dimensional features and others, effectively through multi-view learning and feed adjusting. Our model is fast and generalize well to many other sequences from the cloud computing platform.


\section*{Acknowledgments}
This work was supported by the National Natural Science Foundation of China under grants 61373127, 61772252, the Young Scientists Fund of the National Natural Science Foundation of China under grants 61702242 and the Doctoral Scientific Research Foundation of Liaoning Province under grants 20170520207.
\par The authors would like to thank the anonymous reviewers for the valuable suggestions they provided.

\end{document}